%% file: paper.tex
\documentclass{article}
\usepackage[letterpaper, left=1in, right=1in, top=1in, bottom=1in]{geometry}
\usepackage{natbib}
\bibpunct{(}{)}{;}{a}{,}{,}
\usepackage[colorlinks,citecolor=blue!70!black,linkcolor=blue!70!black,
urlcolor=blue!70!black,breaklinks=true]{hyperref}

\usepackage{url}            %
\usepackage{booktabs}       %
\usepackage{amsfonts}       %
\usepackage{nicefrac}       %
\usepackage{microtype}      %
\usepackage{xcolor}         %

\usepackage{amsmath}
\usepackage{dsfont} 
\usepackage{amssymb}
\usepackage{xspace}
\usepackage{algorithm}
\usepackage[noend]{algorithmic}
\usepackage{amsthm}
\usepackage{listings}
\usepackage{bm}
\usepackage{bbm}
\usepackage{enumitem}
\usepackage{nicefrac}       
\usepackage{mathrsfs} 
\usepackage{inconsolata}
\usepackage{wrapfig}
\usepackage{graphicx}
\graphicspath{{figs/}}
\usepackage{bigints}
\usepackage{transparent}
\usepackage{comment}
\usepackage[nameinlink,capitalize]{cleveref}
\crefname{figure}{Figure}{Figures}
\makeatletter
\AddToHook{cmd/appendix/before}{
\def\cref@section@alias{appendix}
\def\cref@subsection@alias{appendix}
\def\cref@subsubsection@alias{appendix}
}
\makeatother
\usepackage{thmtools}
\usepackage{thm-restate}
\usepackage{nicematrix}
\usepackage{arydshln}
\usepackage{fancyhdr}
\usepackage{mathtools}
\usepackage[scaled=.88]{helvet}
\usepackage{breakcites}
\usepackage{multirow}
\usepackage{subcaption}
\usepackage{caption}
\usepackage{parskip}        %

\input{macros}

\newcommand{\myalgo}{\textsc{RefGRPO}\xspace}  
\newcommand{\grpoplus}{GRPO$^{+}$\xspace}  
\newcommand{\advgrpo}{A^{\mathrm{grpo}}}  
\newcommand{\rawcal}{c}  
\newcommand{\refsc}{s^{\mathrm{ref}}}  
\newcommand{\outcome}{r}  
\newcommand{\refcoeff}{\alpha}  %

\newcommand{\action}{a}
\newcommand{\obs}{o}
\newcommand{\policy}{\pi_\theta}
\newcommand{\accmath}{{\mathsf{Acc}}}
\newcommand{\refaccmath}{{\mathsf{Acc}_{\mathsf{ref}}}}
\newcommand{\chowmath}{{\mathsf{ChowScore}}}
\newcommand{\overmath}{{\mathsf{OverConf}}}             %
\newcommand{\undermath}{{\mathsf{UnderConf}}}           %
\newcommand{\validmath}{{\mathsf{P}_{\mathsf{valid}}}}  %

\newcommand{\acctext}{{$\accmath$}\xspace}
\newcommand{\refacctext}{{$\refaccmath$}\xspace}
\newcommand{\chowtext}{{$\chowmath$}\xspace}

\newcommand{\overtext}{{$\overmath$}\xspace}
\newcommand{\undertext}{{$\undermath$}\xspace}
\newcommand{\validtext}{{$\validmath$}\xspace}

\definecolor{darkblue}{rgb}{0, 0, 0.5}
\hypersetup{colorlinks=true, citecolor=darkblue, linkcolor=darkblue, urlcolor=darkblue}

\title{Closing the Reflection Gap: A Free Calibration Bonus for Agentic RL}

\author{%
  Yinglun Zhu\\
  {\normalsize \texttt{yzhu@ucr.edu}} \\
  {\normalsize University of California, Riverside}
}
\date{}

\begin{document}

\maketitle

\begin{abstract}
\input{sections/abstract}

\end{abstract}

\input{sections/intro.tex}

\input{sections/setting.tex}
\input{sections/method.tex}

\input{sections/experiment.tex}
\input{sections/related.tex}

\input{sections/conclusion.tex}

\bibliography{refs}
\bibliographystyle{plainnat}

\newpage
\appendix

\input{sections/appendix.tex}

\end{document}

%% file: macros.tex
\DeclareMathOperator{\clip}{\mathrm{clip}}

\DeclareMathOperator{\old}{{old}}

\renewcommand{\epsilon}{\varepsilon}

\newcommand{\indic}{\mathbb{I}}

\newtheoremstyle{spaced}
  {6pt}   %
  {0pt}   %
  {\itshape} %
  {}       %
  {\bfseries} %
  {.}      %
  {0.5em}  %
  {}
\theoremstyle{spaced}

\newcommand{\algcommentlight}[1]{\textcolor{blue!70!black}{\transparent{0.5}\small{\texttt{\textbf{//\hspace{2pt}#1}}}}}

\DeclarePairedDelimiterX{\infdiv}[2]{(}{)}{%
  #1\;\delimsize\|\;#2%
}

\newcommand{\mc}[1]{\mathcal{#1}}

\def\ddefloop#1{\ifx\ddefloop#1\else\ddef{#1}\expandafter\ddefloop\fi}
\def\ddef#1{\expandafter\def\csname bb#1\endcsname{\ensuremath{\mathbb{#1}}}}
\ddefloop ABCDEFGHIJKLMNOPQRSTUVWXYZ\ddefloop
\def\ddefloop#1{\ifx\ddefloop#1\else\ddef{#1}\expandafter\ddefloop\fi}
\def\ddef#1{\expandafter\def\csname b#1\endcsname{\ensuremath{\mathbf{#1}}}}
\ddefloop ABCDEFGHIJKLMNOPQRSTUVWXYZ\ddefloop
\def\ddef#1{\expandafter\def\csname sf#1\endcsname{\ensuremath{\mathsf{#1}}}}
\ddefloop ABCDEFGHIJKLMNOPQRSTUVWXYZ\ddefloop
\def\ddef#1{\expandafter\def\csname c#1\endcsname{\ensuremath{\mathcal{#1}}}}
\ddefloop ABCDEFGHIJKLMNOPQRSTUVWXYZ\ddefloop
\def\ddef#1{\expandafter\def\csname h#1\endcsname{\ensuremath{\widehat{#1}}}}
\ddefloop ABCDEFGHIJKLMNOPQRSTUVWXYZ\ddefloop
\def\ddef#1{\expandafter\def\csname hc#1\endcsname{\ensuremath{\widehat{\mathcal{#1}}}}}
\ddefloop ABCDEFGHIJKLMNOPQRSTUVWXYZ\ddefloop
\def\ddef#1{\expandafter\def\csname t#1\endcsname{\ensuremath{\widetilde{#1}}}}
\ddefloop ABCDEFGHIJKLMNOPQRSTUVWXYZ\ddefloop
\def\ddef#1{\expandafter\def\csname tc#1\endcsname{\ensuremath{\widetilde{\mathcal{#1}}}}}
\ddefloop ABCDEFGHIJKLMNOPQRSTUVWXYZ\ddefloop
\def\ddefloop#1{\ifx\ddefloop#1\else\ddef{#1}\expandafter\ddefloop\fi}
\def\ddef#1{\expandafter\def\csname scr#1\endcsname{\ensuremath{\mathscr{#1}}}}
\ddefloop ABCDEFGHIJKLMNOPQRSTUVWXYZ\ddefloop

\let\oldparagraph\paragraph
\renewcommand{\paragraph}[1]{\oldparagraph{#1}}

\renewcommand{\epsilon}{\varepsilon}

\newcommand{\eps}{\epsilon}
\newcommand{\epslow}{\eps_{\text{low}}}
\newcommand{\epshigh}{\eps_{\text{high}}}

\renewcommand{\bigm}[1]{%
  \ifcsname fenced@\string#1\endcsname
    \expandafter\@firstoftwo
  \else
    \expandafter\@secondoftwo
  \fi
  {\expandafter\amsmath@bigm\csname fenced@\string#1\endcsname}%
  {\amsmath@bigm#1}%
}

\newcommand{\DeclareFence}[2]{\@namedef{fenced@\string#1}{#2}}
\makeatother

\makeatletter
\let\save@mathaccent\mathaccent
\newcommand*\if@single[3]{%
  \setbox0\hbox{${\mathaccent"0362{#1}}^H$}%
  \setbox2\hbox{${\mathaccent"0362{\kern0pt#1}}^H$}%
  \ifdim\ht0=\ht2 #3\else #2\fi
  }
\newcommand*\rel@kern[1]{\kern#1\dimexpr\macc@kerna}
\newcommand*\widebar[1]{\@ifnextchar^{{\wide@bar{#1}{0}}}{\wide@bar{#1}{1}}}
\newcommand*\wide@bar[2]{\if@single{#1}{\wide@bar@{#1}{#2}{1}}{\wide@bar@{#1}{#2}{2}}}
\newcommand*\wide@bar@[3]{%
  \begingroup
  \def\mathaccent##1##2{%
    \let\mathaccent\save@mathaccent
    \if#32 \let\macc@nucleus\first@char \fi
    \setbox\z@\hbox{$\macc@style{\macc@nucleus}_{}$}%
    \setbox\tw@\hbox{$\macc@style{\macc@nucleus}{}_{}$}%
    \dimen@\wd\tw@
    \advance\dimen@-\wd\z@
    \divide\dimen@ 3
    \@tempdima\wd\tw@
    \advance\@tempdima-\scriptspace
    \divide\@tempdima 10
    \advance\dimen@-\@tempdima
    \ifdim\dimen@>\z@ \dimen@0pt\fi
    \rel@kern{0.6}\kern-\dimen@
    \if#31
      \overline{\rel@kern{-0.6}\kern\dimen@\macc@nucleus\rel@kern{0.4}\kern\dimen@}%
      \advance\dimen@0.4\dimexpr\macc@kerna
      \let\final@kern#2%
      \ifdim\dimen@<\z@ \let\final@kern1\fi
      \if\final@kern1 \kern-\dimen@\fi
    \else
      \overline{\rel@kern{-0.6}\kern\dimen@#1}%
    \fi
  }%
  \macc@depth\@ne
  \let\math@bgroup\@empty \let\math@egroup\macc@set@skewchar
  \mathsurround\z@ \frozen@everymath{\mathgroup\macc@group\relax}%
  \macc@set@skewchar\relax
  \let\mathaccentV\macc@nested@a
  \if#31
    \macc@nested@a\relax111{#1}%
  \else
    \def\gobble@till@marker##1\endmarker{}%
    \futurelet\first@char\gobble@till@marker#1\endmarker
    \ifcat\noexpand\first@char A\else
      \def\first@char{}%
    \fi
    \macc@nested@a\relax111{\first@char}%
  \fi
  \endgroup
}
\makeatother

%% file: sections/abstract.tex
LLMs are increasingly deployed as agents that interact with external environments and observe feedback such as execution results, error messages, and tool outputs. A well-functioning agent should be able to leverage this feedback to accurately assess its own performance. Yet we find a persistent \emph{reflection gap}: LLM agents tend to mis-assess their own outputs after observing concrete environment feedback---even for questions they correctly answered---and standard RL barely helps due to a credit-assignment mismatch. To close this gap, we propose \myalgo, a simple yet effective fix that augments standard RL algorithms with two key ingredients: a free calibration bonus computed by contrasting the agent's own reflection with the actual outcome (requiring no additional reward model, LLM judge, or external annotation), and a dynamic schedule on its coefficient. Compared to standard RL baselines, our method simultaneously improves reflection calibration (e.g., reduces underconfidence rate $44.4\% \to 7.7\%$) and task accuracy (e.g., $75.1\% \to 76.5\%$) on text-to-SQL across five benchmarks. The resulting calibrated reflection turns the agent into its own verifier grounded in environment feedback, which further enables (i) better self-improvement that uses reflections as pseudo-rewards without outcome supervision, and (ii) more effective test-time selective prediction by committing only to rollouts flagged as correct.

%% file: sections/intro.tex
\begin{figure}[H]
  \centering
  \includegraphics[width=\textwidth]{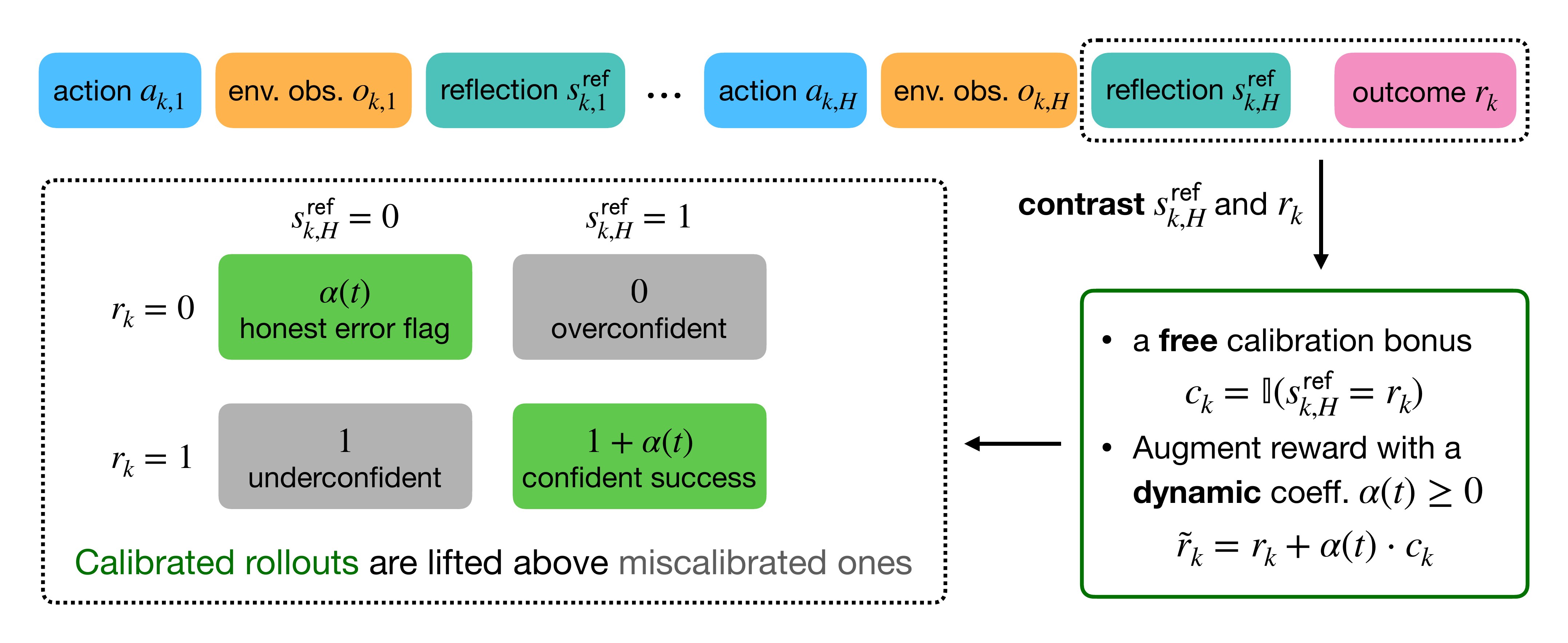}
  \caption{
    High-level overview of our \myalgo algorithm.
  We instruct the agent to reflect on environment feedback and generate a binary reflection score $\refsc \in \{0,1\}$.  
  \myalgo has two key ingredients:
  (i) a \emph{free} calibration bonus $\rawcal_k = \indic(\refsc_{k,H} = \outcome_k)$ computed by contrasting post-feedback reflection with the outcome, requiring no additional reward model, LLM judge, or external annotation; it lifts well-calibrated rollouts above miscalibrated ones, giving honest reflection positive relative advantage regardless of task outcome.
  (ii) a \emph{dynamic} schedule on the calibration coefficient, which enables the model to simultaneously improve reflection calibration and task performance.
    In effect, \myalgo turns the agent into its own verifier grounded in environment feedback.
  }
  \label{fig:workflow}
\end{figure}

\section{Introduction}
\label{sec:intro}

\begin{figure}[t]
\centering
\includegraphics[width=\textwidth]{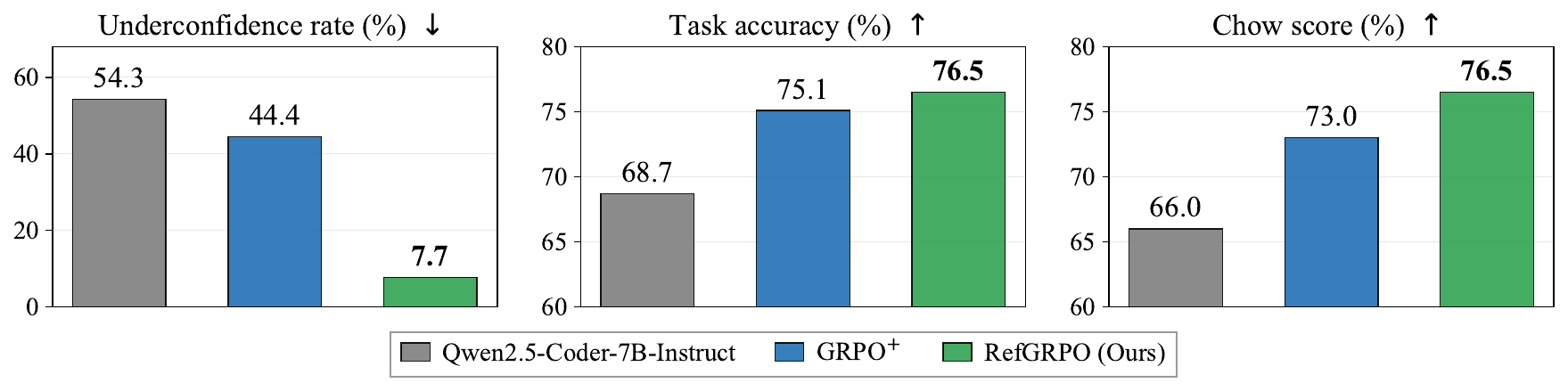}
\caption{Comparison of the base model, a GRPO-style RL baseline, and our \myalgo in the multi-turn setting. 
The model generates reflection scores $\refsc \in \{0,1\}$ based on environment feedback before receiving the outcome reward $r \in \{0,1\}$.
\textbf{(a)} underconfidence rate $\undermath = \bbP(\outcome= 1 \mid \refsc= 0)$, the fraction of
self-flagged errors that are actually correct (\emph{lower is better}). \textbf{(b)} task accuracy $\accmath = \bbP(\outcome = 1)$ (\emph{higher is better}). \textbf{(c)} a unified metric \chowtext for task accuracy and reflection calibration (\emph{higher is better}).
\looseness=-1
}
\label{fig:underconf}
\end{figure}

LLM agents acting in an environment receive \emph{environment feedback}, e.g., execution results, error messages, or tool outputs, after each action \citep{jin2025searchr, cao2025skyrl, zhang2026the}. This feedback is concrete evidence about whether an action succeeded, yet agents typically consume it only as context for choosing the next action, rarely explicitly using it to judge whether the task itself has been correctly solved.

In this work, we explicitly instruct the agent to \emph{reflect} on the environment feedback and produce a \emph{binary score} indicating whether its answer is correct---a post-feedback self-assessment grounded in evidence it has already observed. This casts reflection as a prediction problem: an agent that can accurately predict its own correctness from the feedback it has seen demonstrates good understanding of the consequences of its actions. Post-feedback reflection quality is therefore a direct measure of how well the agent comprehends what it has done and how the environment evolves, i.e., the quality of its \emph{implicit} world model \citep{ha2018world}.
\looseness=-1

Crucially, this is different from existing work on LLM self-assessment \citep{kadavath2022language, tian2023just, xiong2024can, tao2024trust, leng2025taming, bani-harouni2026rewarding, damani2026beyond}, 
where a model expresses its confidence in a self-generated answer \emph{solely} based on its own generation, \emph{without environment feedback}.
In our setting, the agent reflects on its own actions \emph{after} observing concrete environment feedback---strictly more information than the self-generated answer alone. 
Interpreting evidence one has already seen should be easier than forecasting correctness blind, so we would expect agents---especially RL-trained agents with strong reasoning capabilities---to excel at it.

Yet they do not. 
Beyond the well-documented overconfidence of LLMs \citep{xiong2024can, leng2025taming}, we find a more surprising failure: LLM agents exhibit significant \emph{underconfidence}---flagging correct answers as wrong---even after observing concrete environment feedback, and standard RL recipes fail to fix it.
On a multi-turn text-to-SQL setup, a Qwen2.5-Coder-7B-Instruct base is badly underconfident: $54.3\%$ of the answers it flags as wrong are actually correct (\cref{fig:underconf}(a)).
Training that base with a GRPO-style algorithm \citep{shao2024deepseekmath, guo2025deepseek, yu2025dapo, liu2025understanding, he2025justrl} only nudges this rate down to $44.4\%$ even though it lifts task accuracy substantially. 
Standard RL recipes make the model a stronger task solver, but leave its post-feedback reflection quality broken---a persistent failure we call the \emph{reflection gap}. The cause is structural: outcome-only RL assigns advantage purely based on the outcome, so a rollout with an incorrect outcome but an honest error flag receives negative advantage, training the agent to suppress correct error flags rather than reward them; see \cref{sec:methods:problem} for a detailed discussion.

To close the reflection gap, we propose a new algorithm \myalgo that extends outcome-only RL with two key ingredients (\cref{fig:workflow}).
(i) A \emph{free} calibration bonus: since both reflection $\refsc$ and the outcome $\outcome$ are already available during RL training, we add a calibration bonus $\rawcal = \indic(\refsc = \outcome)$ to the outcome reward before group normalization. This lifts well-calibrated rollouts above miscalibrated ones, giving honest reflection positive relative advantage regardless of task outcome.
(ii) A dynamic schedule on the calibration coefficient $\refcoeff(t) \geq 0$, which starts with a relatively large value to front-load calibration early and then decays the coefficient to let the
model focus on task performance while largely retaining the calibration ability.
As shown in \cref{fig:underconf}, compared to a GRPO-style baseline, \myalgo significantly reduces the underconfidence rate $44.4\% \to 7.7\%$, while simultaneously improving the task accuracy $75.1\% \to 76.5\%$.
We also introduce the \chowtext from statistical learning \citep{chow1957optimum, chow1970optimum} to the agentic setting as a unified metric for task accuracy and reflection calibration;
\myalgo improves \chowtext $73.0\% \to 76.5\%$.
\looseness=-1

\paragraph{Contributions.}
We make the following main contributions:
\begin{enumerate}[leftmargin=*]
  \item \textbf{Post-feedback reflection is broken, and outcome-only RL barely fixes it.} 
  We show LLM agents tend to mis-assess their own outputs even after observing concrete environment feedback, with underconfidence rates above $44\%$ in the multi-turn setting (\cref{fig:underconf}).
  We propose metrics to quantify the reflection gap (\cref{sec:setting:metrics}), and trace it to a credit-assignment mismatch problem (\cref{sec:methods:problem}).
  
  \item \textbf{Augmenting RL with a free calibration bonus.} We augment standard RL recipes with (i) a free calibration bonus computed from the contrast between post-feedback reflection and outcome, and (ii) a dynamic schedule on its coefficient (\cref{sec:methods:algorithm}).
  As shown in \cref{fig:underconf,sec:exp:main}, our algorithm simultaneously improves reflection calibration (e.g., reduces underconfidence rate $44.4\% \to 7.7\%$) and task accuracy (e.g., $75.1\% \to 76.5\%$), lifting the unified \chowtext metric $73.0\% \to 76.5\%$.

\item \textbf{Calibration enables better self-improvement and selective prediction.} The resulting calibrated reflection turns the agent into its own verifier---one \emph{grounded in environment feedback} rather than pure self-assessment. We show that this further enables (i) better self-improvement that uses reflections as pseudo-rewards \emph{without} outcome supervision (\cref{sec:exp:self}), and (ii) more effective test-time selective prediction by committing only to rollouts flagged as correct (\cref{sec:exp:select}).

\end{enumerate}

%% file: sections/setting.tex
\section{Problem Setting}
\label{sec:setting}

We formalize the interaction between agents and environments, and introduce 
the metrics to evaluate the quality of an agent's reflection  based on both
its own actions and the environment feedback.

\paragraph{The interaction framework.}
\label{sec:setting:refenv}
We study an environment in which the agent assesses its action
\emph{after observing concrete environment feedback}. Each turn 
proceeds in three stages (see \cref{fig:workflow}, top):
\begin{enumerate}[leftmargin=*]
  \item \textbf{Action.} The agent takes an action $\action$
  (e.g., a SQL query) after step-by-step reasoning.
  \item \textbf{Observation.} The environment executes the action $\action$ and
  returns an observation $\obs$ (e.g., query results, error messages, or
  tool outputs).
  \item \textbf{Reflection.} The agent reflects on \emph{both} 
  the action $\action$ and the observation $\obs$ and
  generates a binary reflection score
  $\refsc \in \{0, 1\}$ indicating whether it believes the
  action successfully completed the task (i.e., $\refsc = 1$) or not (i.e., $\refsc= 0$).
\end{enumerate}
After the final turn, the environment provides a binary outcome reward
$\outcome \in \{0, 1\}$ based on the correctness of the action; the outcome
reward is used for RL training but is \emph{not}
 revealed to the agent in the interaction. 

\paragraph{Single-turn and multi-turn settings.}
We study both single-turn and multi-turn interactions.
In the \emph{single-turn} setting, the agent takes an action,
observes the environment feedback, and 
conducts the reflection.\footnote{Our
single-turn setting differs from the non-agentic single-turn setup
(e.g., \citet{damani2026beyond}): the agent reflects on its action
together with the resulting environment feedback, rather than on
its self-generated answer alone.}
In the \emph{multi-turn} setting, the agent interacts with the environment
up to $H$ turns.
The agent can commit to a final action early and terminate the episode if it is confident about the result.
At each turn,
we instruct the agent to reflect on \emph{all} preceding actions and observations and generate a reflection score;
we therefore use the last reflection score $\refsc_H$
as the overall self-assessment for the episode.

\paragraph{Evaluation metrics.}
\label{sec:setting:metrics}
We measure task performance and reflection quality with the following metrics.
\emph{Task accuracy} $\accmath = \bbP(\outcome = 1)$ measures whether
the agent's answer is correct.
\emph{Reflection accuracy} $\refaccmath = \bbP(\refsc = \outcome)$
measures the agreement between the agent's reflection and
the actual outcome. 
We also consider two directional miscalibration rates to decompose
the failure modes when $\refsc \neq \outcome$:
\emph{overconfidence rate} $\overmath = \bbP(\outcome = 0 \mid
\refsc = 1)$, the fraction of answers it flags as correct that are actually wrong; and \emph{underconfidence rate}
$\undermath = \bbP(\outcome = 1 \mid \refsc = 0)$, the fraction of answers it flags as wrong that are actually correct (lower is better for both).

\paragraph{A unified metric: Chow score.}
We introduce the \chowtext\ \citep{chow1957optimum, chow1970optimum} from statistical learning to the agentic setting as a unified metric for task accuracy and \emph{reflection calibration}---the agent's ability to know when it has solved the task. 
Specifically, 
the \chowtext 
scores commits
($\refsc =1$) by correctness and self-flagged errors ($\refsc = 0$) by a fixed
credit $\beta \in [0, 1)$:
\looseness=-1
\begin{equation*}
  \chowmath_\beta = \bbP(\refsc = 1, \outcome = 1)
  + \beta \cdot \bbP(\refsc = 0).
\end{equation*}
$\chowmath_\beta = \accmath$ for an always-commit agent and exceeds \acctext whenever the agent's error flags are informative; $\beta$ controls how much we credit honest error detection (which can be interpreted as abstention).
We treat \chowtext as the \emph{main metric} since it captures both task accuracy and reflection calibration.
We set $\beta = 0.1$  by default and report a sweep over $\beta \in [0, 0.5]$ in the ablations (\cref{sec:exp:ablations}).

%% file: sections/method.tex
\section{Methods}
\label{sec:methods}

\begin{figure}[t]
  \centering
  \includegraphics[width=\textwidth]{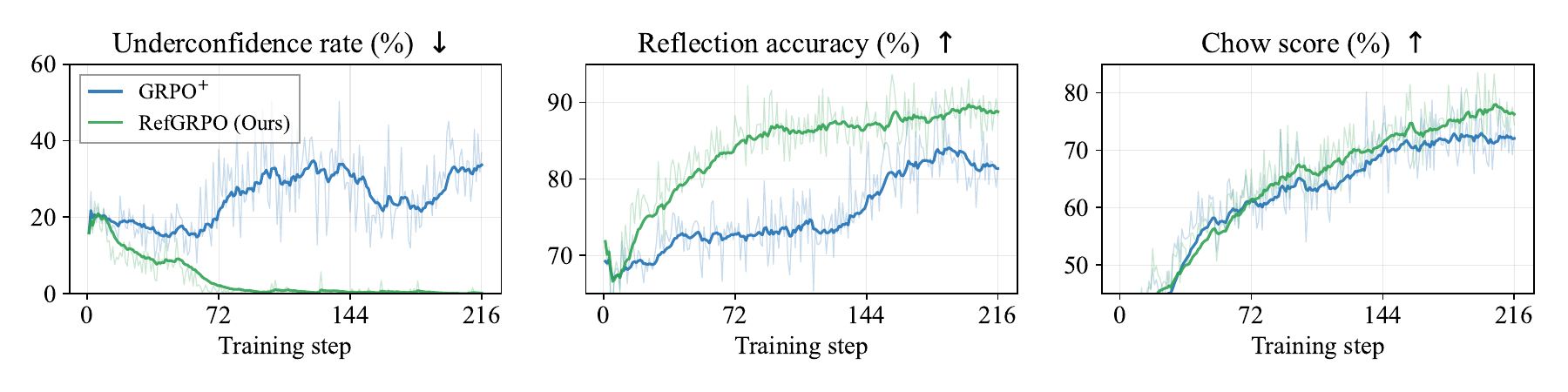}
  \caption{Training curves for \grpoplus and \myalgo (Ours) in the
  single-turn setting using Qwen2.5-Coder-3B-Instruct as the base model. \textbf{Left:}
  underconfidence rate (\emph{lower is better});
  \textbf{Middle:} reflection accuracy (\emph{higher is better}); \textbf{Right:}
  \chowtext at $\beta{=}0.1$ (\emph{higher is better}).}
  \label{fig:training_curves}
\end{figure}

We analyze why outcome-only RL leads to uncalibrated models (\cref{sec:methods:problem}) and propose a new algorithm that closes the reflection gap (\cref{sec:methods:algorithm}).

\subsection{Limitations of Outcome-Only RL}
\label{sec:methods:problem}

\paragraph{Background: GRPO.}
Group Relative Policy Optimization
(GRPO; \citealt{shao2024deepseekmath, guo2025deepseek}) generates $G$ rollouts
$\{\tau_k\}_{k=1}^G$ for each prompt $q \sim \mc{D}$ and updates the
policy via the following clipped surrogate objective \citep{schulman2017proximal, yu2025dapo}:
\begin{equation}
  J_{\text{\grpoplus}}(\theta) = \bbE_{q,\,\{\tau_k\}} 
  \left[
    \tfrac{1}{N}\sum_{k=1}^{G}\sum_{t=1}^{|\tau_k|}
    \min\bigl(
      \rho_{k,t}\,\advgrpo_k,\;
      \clip(\rho_{k,t}, 1- \epslow, 1 +\epshigh)\,\advgrpo_k
    \bigr)
  \right],
  \label{eq:grpo_loss}
\end{equation}
where $\rho_{k,t} = \policy (\action_{k,t}\mid h_{k,t}) /
\pi_{\theta_{\old}}(\action_{k,t} \mid h_{k,t})$ is the importance
ratio and the advantage $\advgrpo_k =
(\outcome_k - \mu) / (\sigma + \eps)$ is normalized with respect to the group mean $\mu$
and standard deviation $\sigma$ over outcome rewards $\{\outcome_j\}_{j=1}^G$. 
We additionally (i) normalize by total number of tokens $N = \sum_k |\tau_k|$ 
to remove length bias, (ii) apply the asymmetric $\clip(\rho, 1-\epslow, 1+\epshigh)$ to encourage exploration and prevent entropy collapse, and (iii) drop the KL divergence term to avoid over-constraining the policy and allow learning to be driven more directly by verifiable rewards \citep{yu2025dapo,liu2025understanding, he2025justrl}.
We denote the resulting objective in \cref{eq:grpo_loss} as \grpoplus to distinguish it from the original GRPO \citep{shao2024deepseekmath}.
\looseness=-1

\paragraph{Credit assignment mismatch.}
\label{sec:methods:mismatch}
The advantage $\advgrpo_k$ in \cref{eq:grpo_loss} depends only on the outcome $\outcome_k$, so the
gradient is purely driven by task correctness.
On rollouts where the agent self-flags its final action as wrong
($\refsc_{k,H}=0$), the signal can be misleading for 
reflection quality:
\begin{itemize}[leftmargin=*]
  \item \textbf{Honest error flag ($\outcome_k=0$, $\refsc_{k,H} =0$):}
  even though the reflection is correct, the 
  negative advantage $\advgrpo_k < 0$ induced by the outcome trains the agent to \emph{stop flagging genuine errors}.
  \item \textbf{Underconfident error flag ($\outcome_k=1$,
  $\refsc_{k,H} =0$):} the positive advantage $\advgrpo_k > 0$ induced by the outcome trains
  the agent to \emph{wrongly doubt its success}.
\end{itemize}

Both cases harm the agent's post-feedback reflection quality.
As shown in \cref{fig:training_curves}, under standard outcome-only RL (e.g., \grpoplus), \undertext\ drifts upward as training proceeds, while \refacctext\ and \chowtext\ lag far behind those of our method (\myalgo, introduced in \cref{sec:methods:algorithm}).

\begin{algorithm}[t]
\caption{\myalgo: GRPO Augmented with a Free Calibration Bonus and a Dynamic Schedule}
\renewcommand{\algorithmicrequire}{\textbf{Input:}}
\renewcommand{\algorithmicensure}{\textbf{Output:}}
\label{alg:refgrpo}
\begin{algorithmic}[1]
\REQUIRE Schedule $\{\refcoeff(t)\}_{t \geq 0}$ with $\refcoeff(t) \geq 0$
\medskip
\FOR{each training step $t$}
  \STATE Get $\refcoeff \gets \refcoeff(t)$ from the schedule
  \FOR{each rollout $k = 1, \dots, G$}
    \STATE Extract the last reflection score $\refsc_{k,H}$
    \STATE Get $\rawcal_k \gets \indic(\refsc_{k,H} = \outcome_k)$
      \algcommentlight{Calibration bonus}
    \STATE Augment the reward with $\tilde{\outcome}_k \gets \outcome_k +
      \refcoeff \cdot \rawcal_k$
      \algcommentlight{Augmented reward}
  \ENDFOR
  \STATE
  Compute 
    $\advgrpo_k \gets
      \dfrac{\tilde{\outcome}_k - \mu}{\sigma + \eps}$
   with 
    $\mu = \tfrac{1}{G}\sum_{j=1}^{G} \tilde{\outcome}_j$
    and
    $\sigma^2 = \tfrac{1}{G}\sum_{j=1}^{G} (\tilde{\outcome}_j - \mu)^2$
    \algcommentlight{GRPO advantage}
  \STATE Update policy with $\{\advgrpo_k\}$ via the RL objective
    in \cref{eq:grpo_loss}
\ENDFOR
\end{algorithmic}
\end{algorithm}

\subsection{\myalgo: Free Calibration from the Reflection-Outcome Contrast}
\label{sec:methods:algorithm}

We introduce a new algorithm \myalgo (\cref{alg:refgrpo}) to resolve the credit-assignment mismatch in outcome-only RL, thereby closing the reflection gap.
\myalgo has two key ingredients: (i) a free calibration bonus computed by contrasting the agent's reflection with the actual outcome, and (ii) a dynamic schedule on the calibration coefficient.

\paragraph{A free calibration bonus from the reflection-outcome contrast.}
The key observation is that, for each rollout $k$, the agent's
post-feedback reflection $\refsc_{k,H}$ and the outcome
reward $\outcome_k$ are both available during training, and
\emph{the contrast between them} can be used to compute a 
calibration signal.
Specifically, 
we compute the binary calibration bonus 
$\rawcal_k = \indic(\refsc_{k,H} = \outcome_k)$ 
and augment the outcome reward $\outcome_k$ with this calibration bonus (with coefficient $\refcoeff(t) \geq 0$ at training step $t$):
\begin{equation*}
  \tilde{\outcome}_k(t) = \outcome_k + \refcoeff(t) \cdot \rawcal_k.
\end{equation*}
This augmentation assigns positive relative advantage to correct post-feedback reflection \emph{independently} of the task outcome.
As shown in \cref{fig:workflow}, no matter whether the task succeeds or not, a well-calibrated rollout receives higher reward than a miscalibrated one with the same outcome, which pushes the agent toward better-calibrated reflection.
\cref{fig:training_curves} further corroborates this effect: compared to \grpoplus, \myalgo significantly reduces \undertext and improves \refacctext, while achieving higher \chowtext.

Notably, the calibration bonus is \emph{free}: it is computed by contrasting the agent's own reflection with the actual outcome, requiring no additional reward model, LLM judge, or external annotation.

\paragraph{A dynamic schedule on the calibration bonus coefficient.}
\label{sec:methods:schedule}

While the augmented reward $\tilde{\outcome}_k$ helps the agent improve its reflection calibration, 
it can slightly reduce task accuracy as part of the signal is reserved for calibration quality rather than correctness.
To balance the two, we introduce a dynamic schedule on the calibration bonus coefficient $\refcoeff(t)\geq 0$ that decays the calibration bonus as training proceeds.
We instantiate $\refcoeff(t)$ as a
simple \emph{two-stage} schedule:
\begin{equation*}
  \refcoeff(t) =
  \begin{cases}
    \refcoeff_0 & t \leq  \gamma \cdot T, \\
    \refcoeff_1 & t > \gamma \cdot T,
  \end{cases}
  \qquad \refcoeff_1 < \refcoeff_0,
\end{equation*}
where $T$ is the total number of training steps. The relatively
larger coefficient $\refcoeff_0$ in the first $\gamma \cdot T$ steps front-loads calibration; the
smaller coefficient $\refcoeff_1$ in the remaining $(1-\gamma) \cdot T$ steps lets the 
model focus on task performance while largely retaining the calibration ability built up in the first stage.
By default we set $\refcoeff_0 = 0.1$, $\refcoeff_1 = 0$, and $\gamma = 2/3$.
As shown in \cref{fig:underconf,tab:main_single,tab:main_multi}, \myalgo with this schedule improves both calibration quality and task accuracy compared to outcome-only RL baselines.
We provide ablations of the dynamic schedule in \cref{sec:exp:ablations}.
\looseness=-1

%% file: sections/experiment.tex
\section{Experiments}
\label{sec:experiments}

\begin{table}[t]
\centering
\caption{Single-turn results across five metrics: task accuracy, reflection accuracy, overconfidence rate, underconfidence rate, and Chow score. All values are percentages. Best results are highlighted in \textbf{bold}. Results are averaged across 5 benchmarks; per-dataset breakdowns are provided in \cref{sec:appendix:per_dataset}.}
\label{tab:main_single}
\small
\setlength{\tabcolsep}{5pt}
\begin{tabular}{lccccc}
\toprule
Method & $\accmath$ $\uparrow$ & $\refaccmath$ $\uparrow$
  & $\overmath$ $\downarrow$ & $\undermath$ $\downarrow$
  & $\chowmath_{0.1}$ $\uparrow$ \\
\midrule
\multicolumn{6}{l}{\emph{Llama-3.2-3B-Instruct}} \\
Base               & 52.1 & 65.1 & 37.5 & 31.4 & 49.0 \\
\grpoplus          & 62.9 & 74.5 & 26.5 & 30.9 & 58.4 \\
\myalgo (Ours)     & \textbf{64.0} & \textbf{76.0} & \textbf{26.3} & \textbf{23.5} & \textbf{62.2} \\
\midrule
\multicolumn{6}{l}{\emph{Qwen2.5-Coder-3B-Instruct}} \\
Base               & 56.8 & 71.4 & 31.4 & 25.2 & 53.8 \\
\grpoplus          & 69.4 & 76.6 & 24.6 & 23.7 & 68.5 \\
\myalgo (Ours)     & \textbf{70.2} & \textbf{79.7} & \textbf{23.1} & \textbf{1.3} & \textbf{71.1} \\
\bottomrule
\end{tabular}
\end{table}

\subsection{Experimental Setup}
\label{sec:exp:setup}

We evaluate on text-to-SQL, a widely-studied agentic environment with
verifiable rewards \citep{yu2018spider, li2023can, gao2024preview,
li2025omnisql, pourreza2025reasoningsql, sqlr1}. It provides
unambiguous binary outcomes (execution correctness) and concrete environment
feedback (SQL query results or error messages) at a difficulty level suited to open-source models.
\looseness=-1

\paragraph{Data and models.}
We train on 4,660 text-to-SQL problems drawn from
the training set of Spider \citep{yu2018spider} and
OmniSQL \citep{li2025omnisql}, and evaluate on five standard benchmarks: Spider-Dev \citep{yu2018spider}, Spider-Domain Knowledge (DK; \citealt{gan2021exploring}), Spider-Realistic \citep{deng2021structure}, Spider-Test \citep{yu2018spider}, and Bird-Dev \citep{li2023can}. 
We provide the average results across five benchmarks in the main content and provide per-dataset breakdowns in the appendix (\cref{app:additional_results}).
We experiment with three instruction-tuned models across two scales:
Qwen2.5-Coder-3B/7B-Instruct \citep{hui2024qwen2} and
Llama-3.2-3B-Instruct \citep{grattafiori2024llama}.\footnote{We use instruction-tuned rather than RL-trained models as our base to avoid potential failure modes induced by standard RL training (\cref{sec:methods:mismatch}).} 

\paragraph{Baselines.}
We use \grpoplus (\cref{eq:grpo_loss}) as the outcome-only RL baseline; it augments the original GRPO objective \citep{shao2024deepseekmath} with DAPO-style improvements \citep{yu2025dapo}---asymmetric clipping, token-mean normalization, and the removal of the KL divergence term.
Our \myalgo (\cref{alg:refgrpo}) further augments \grpoplus with a free calibration bonus and a dynamic schedule on its coefficient (\cref{sec:methods:schedule}).
We train both \grpoplus and \myalgo for 6 epochs, using the same hyperparameters detailed in \cref{app:implementation}.
We
additionally benchmark against two open source 7B SQL specialists:
OmniSQL-7B (SFT with 2.5M CoT data; \citealt{li2025omnisql}) and
SQL-R1-7B (GRPO with outcome reward; \citealt{sqlr1}).

\paragraph{Metrics.}
We report five metrics for both single- and multi-turn (up to 6 turns) results: 
task accuracy \acctext,
reflection accuracy \refacctext, the two miscalibration rates \overtext\ and
\undertext, and the unified metric \chowtext.\footnote{To reduce clutter, in this section we drop the \% symbol
after reported results whenever context makes the unit unambiguous.} 
We consider \chowtext as the \emph{main metric} that captures both task accuracy and reflection calibration (\cref{sec:setting:metrics}).
Following prior work \citep{li2025omnisql, sqlr1}, the main results are reported with greedy decoding; in \cref{sec:exp:select}, we report aggregated results with $k = 8$ samples at temperature $0.6$.
We default to 
$\chowmath_{0.1}$ and provide a sweep over error-flag credits
$\beta \in [0, 0.5]$ in the ablations (\cref{sec:exp:ablations}),
where we also evaluate error detection via other metrics such as precision and
recall.

\subsection{Main Results}
\label{sec:exp:main}

\paragraph{Single-turn results.} \Cref{tab:main_single} reports results in the single-turn setting for two base models (Llama-3.2-3B-Instruct and Qwen2.5-Coder-3B-Instruct). \myalgo dominates \grpoplus on all five metrics across both base models---most notably improving \refacctext\ (Qwen: \grpoplus $76.6 \to $ \myalgo $79.7$; Llama: \grpoplus $74.5 \to$ \myalgo $76.0$) and reducing \undertext\ (Qwen: \grpoplus $23.7 \to$ \myalgo $1.3$; Llama: \grpoplus $30.9 \to$ \myalgo $23.5$). Compared to the base, \grpoplus only marginally reduces \undertext\ (Qwen: Base $25.2 \to$ \grpoplus $23.7$; Llama: Base $31.4 \to$ \grpoplus $30.9$), while \myalgo reduces it substantially (Qwen: Base $25.2 \to$ \myalgo $1.3$; Llama: Base $31.4 \to$ \myalgo $23.5$). The large improvement in reflection quality confirms that the free calibration bonus directly addresses the credit-assignment mismatch in outcome-only RL (\cref{sec:methods:mismatch}). \textbf{Overall, these effects translate into substantial \chowtext\ gains at both bases} (Qwen: \grpoplus $68.5 \to$ \myalgo $71.1$; Llama: \grpoplus $58.4 \to $ \myalgo $62.2$), the main metric that captures both task accuracy and reflection calibration.

\begin{table}[t]
\centering
\caption{Multi-turn results across five metrics: task accuracy, reflection accuracy, overconfidence rate, underconfidence rate, and Chow score. All values are percentages. Best results are highlighted in \textbf{bold}. Results are averaged across 5 benchmarks; per-dataset breakdowns are provided in \cref{sec:appendix:per_dataset}.}
\label{tab:main_multi}
\small
\setlength{\tabcolsep}{5pt}
\begin{tabular}{lccccc}
\toprule
Method & $\accmath$ $\uparrow$ & $\refaccmath$ $\uparrow$
  & $\overmath$ $\downarrow$ & $\undermath$ $\downarrow$
  & $\chowmath_{0.1}$ $\uparrow$ \\
\midrule
\multicolumn{6}{l}{\emph{Qwen2.5-Coder-3B-Instruct}} \\
Base               & 57.2 & 70.7 & 32.8 & 18.7 & 55.8 \\
\grpoplus          & 72.5 & 75.8 & \textbf{24.0} & 36.5 & 70.8 \\
\myalgo (Ours)     & \textbf{72.9} & \textbf{76.3} & 24.8 & \textbf{2.2} & \textbf{73.1} \\
\midrule
\multicolumn{6}{l}{\emph{Qwen2.5-Coder-7B-Instruct}} \\
Base               & 68.7 & 73.1 & 25.0 & 54.3 & 66.0 \\
\grpoplus          & 75.1 & 76.2 & 22.8 & 44.4 & 73.0 \\
\myalgo (Ours)     & \textbf{76.5} & \textbf{77.8} & \textbf{22.4} & \textbf{7.7} & \textbf{76.5} \\
\bottomrule
\end{tabular}
\end{table}

\paragraph{Multi-turn results.}
The multi-turn setting tests long-horizon agentic performance with error correction.
\Cref{tab:main_multi} reports results across five metrics at both 3B and 7B scales.
The results match the pattern in the single-turn setting: 
\myalgo outperforms \grpoplus across metrics, especially in 
improving \refacctext (Qwen-7B: \grpoplus $76.2 \to$ \myalgo $77.8$) and reducing \undertext (Qwen-7B: \grpoplus $44.4 \to$ \myalgo $7.7$; Qwen-3B: \grpoplus $36.5 \to$ \myalgo $2.2$).
On Qwen-3B, \grpoplus achieves slightly better \overtext (\grpoplus $24.0$ vs. \myalgo $24.8$) but at the cost of worsening \undertext over the base model (Base $18.7 \to$ \grpoplus $36.5$);
\myalgo instead reduces both \overtext and \undertext, with especially substantial improvement in \undertext (Base $18.7 \to$ \myalgo $2.2$).
On Qwen-7B, \myalgo also meaningfully lifts \acctext (\grpoplus $75.1 \to$ \myalgo $76.5$).
\textbf{Overall, these effects translate into substantial \chowtext gains at both scales} (Qwen-7B: \grpoplus $73.0 \to$ \myalgo $76.5$; Qwen-3B: \grpoplus $70.8 \to$ \myalgo $73.1$), the main metric that captures both task accuracy and reflection calibration.

\begin{wrapfigure}[15]{r}{0.37\textwidth}
\vspace{-12pt}
\centering
\includegraphics[width=\linewidth]{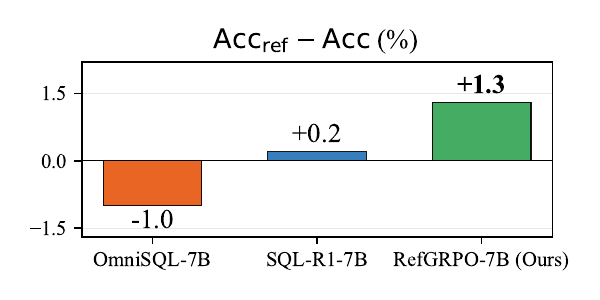}
\caption{Comparison of $\refaccmath - \accmath$ across three 7B models. \emph{Greater than zero indicates the reflection is informative.} Best result is highlighted in \textbf{bold}. Results are averaged across 5 benchmarks; per-dataset breakdowns are provided in \cref{sec:appendix:per_dataset_external}.}
\label{fig:external_comparison}
\vspace{-10pt}
\end{wrapfigure}

\paragraph{Comparison with other 7B SQL specialists.}
We compare our 7B model against two open-weight 7B SQL specialists:
OmniSQL-7B (SFT with 2.5M CoT data; \citealt{li2025omnisql}) and
SQL-R1-7B (GRPO with outcome reward; \citealt{sqlr1}).
Since these specialists use different training data and hyperparameters,
we focus on the calibration delta $\Delta = \refaccmath - \accmath$, 
which isolates the reflection contribution beyond raw task accuracy: a trivial ``always-commit'' model has $\Delta = 0$, while a better-calibrated model achieves $\Delta > 0$ by correctly flagging some of its own errors. As shown in \cref{fig:external_comparison}, \myalgo achieves the largest $\Delta$ at $+1.3$; SQL-R1 is barely positive ($+0.2$); and OmniSQL is even \emph{negative} ($-1.0$), meaning its reflection is not informative---the model can answer some questions correctly but cannot tell whether its own answers are right, even with environment feedback (e.g., SQL query results).

\begin{figure}[t]
  \centering
  \includegraphics[width=\textwidth]{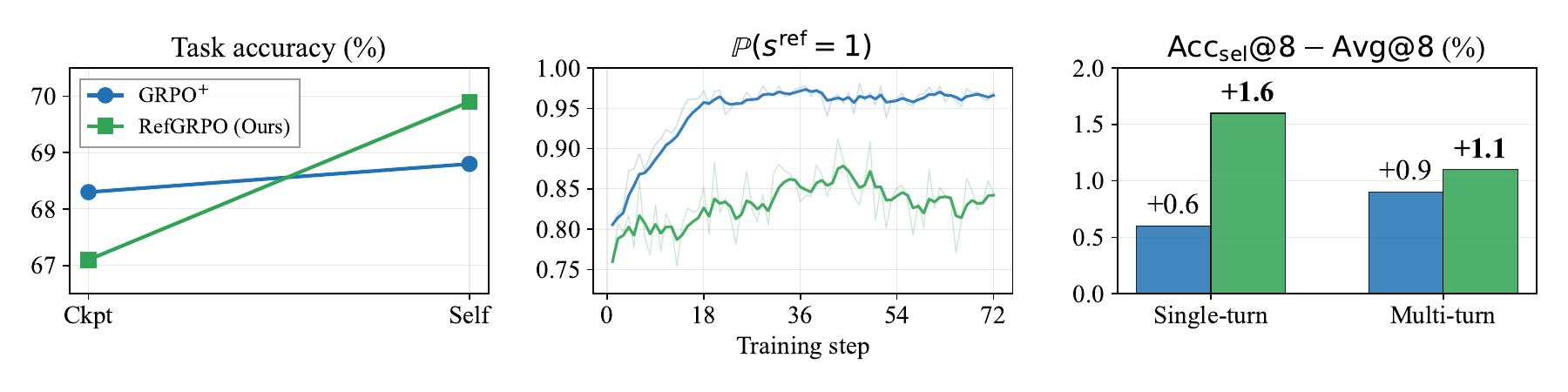}
  \caption{Self-improvement and selective prediction results using 
  Qwen2.5-Coder-3B-Instruct as the base model (averaged across 5 benchmarks; per-dataset breakdowns are provided in \cref{sec:appendix:per_dataset_self,sec:appendix:per_dataset_select}).
  \textbf{(a)} Task accuracy before (Ckpt) and after
  (Self) self-improvement in the single-turn setting.
  \textbf{(b)} Commit rate $\bbP(\refsc= 1)$ during single-turn
  self-improvement training.
  \textbf{(c)} Selective-prediction lift
  $\selaccmath@8 - \mathrm{Avg}@8$ for \grpoplus\ vs.\ \myalgo\ in
  single-turn and multi-turn settings.
  }
  \label{fig:extension}
\end{figure}

\subsection{\myalgo Enables Verifier-Free Self-Improvement}
\label{sec:exp:self}
Calibration turns the agent into its own verifier: its calibrated reflection becomes an informative reward signal, so we can continue RL training using reflection scores as pseudo-rewards \emph{without} any outcome supervision.
To test this, we run a self-supervised RL phase
from \grpoplus and \myalgo checkpoints taken before performance plateaus, with the outcome reward replaced by the agent's reflection
score $\refsc$ (no calibration bonus, $\refcoeff = 0$); reflection
tokens are masked so only task tokens receive gradients. 
We run this phase for 2 epochs with the same hyperparameters as the original training run.

\paragraph{Results.}
\cref{fig:extension}(a) shows the key result:
starting from a \myalgo checkpoint, self-improvement lifts task
accuracy by $+2.8$ points ($67.1 \to 69.9$), while starting from a
\grpoplus checkpoint yields only $+0.5$ points ($68.3 \to 68.8$).
Despite starting from a lower-accuracy checkpoint, the \myalgo trajectory ends with 
higher final task accuracy than \grpoplus (\grpoplus $68.8 \to$ \myalgo $69.9$).

\cref{fig:extension}(b) explains why. During self-improvement, the \grpoplus checkpoint's commit rate $\bbP(\refsc{=}1)$ rapidly saturates near $1.0$: the model asserts confidence on nearly every rollout, so almost all self-rewards collapse to $1$ regardless of correctness. 
This creates two problems: (i) the pseudo-rewards are systematically miscalibrated---since the actual task accuracy is below $70$ (\cref{fig:extension}(a)), the agent rewards itself for many incorrect rollouts; and (ii) \grpoplus's within-group normalization on near-constant rewards produces near-zero advantages, effectively halting learning. The \myalgo\ checkpoint, by contrast, retains a more calibrated commit rate of around $0.83$, so the pseudo-reward inherits the informativeness of $\refsc$ induced by the calibration bonus, and continues to drive policy improvement.

\subsection{\myalgo Enables Better Test-Time Selective Prediction}
\label{sec:exp:select}

In test-time scaling with $k$ rollouts per question, the agent can act as its own verifier, using its reflection $\refsc$ to select which rollouts to commit to.
We measure this with 
\emph{selective prediction accuracy} $\selaccmath@k$, the
expected correctness over the committed set
$C(q) = \{i : \refsc_i = 1\}$:
\begin{equation*}
  \selaccmath@k = \bbE_q \left[\frac{\sum_{i=1}^{k} \indic(\refsc_i=1)\,\outcome_i}
                        {\max\bigl(1,\,|C(q)|\bigr)}\right].
\end{equation*}
On each question we average correctness over the rollouts the agent commits to ($\refsc_i=1$); if it commits to none of the $k$ samples, the question scores $0$. An always-commit agent recovers $\selaccmath@k = \mathrm{Avg}@k$, while a well-calibrated agent achieves $\selaccmath@k > \mathrm{Avg}@k$ by selectively committing to correct rollouts.

\paragraph{Results.}
\cref{fig:extension}(c) reports the selective-prediction \emph{lift}
$\selaccmath@k - \mathrm{Avg}@k$ for \grpoplus and \myalgo in both the
single-turn and multi-turn settings. 
\myalgo
attains the higher lift in \emph{both} regimes: \grpoplus $+0.6 \to$ \myalgo $+1.6$ 
in single-turn, and \grpoplus $+0.9 \to$ \myalgo $+1.1$ in multi-turn---a meaningful improvement especially in the single-turn setting.

\subsection{Analyses and Ablations}
\label{sec:exp:ablations}

We conduct additional analyses and ablations using 
Qwen2.5-Coder-3B-Instruct, reporting averaged results across the
5 evaluation benchmarks.

\begin{figure}[t]
  \centering
  \includegraphics[width=\textwidth]{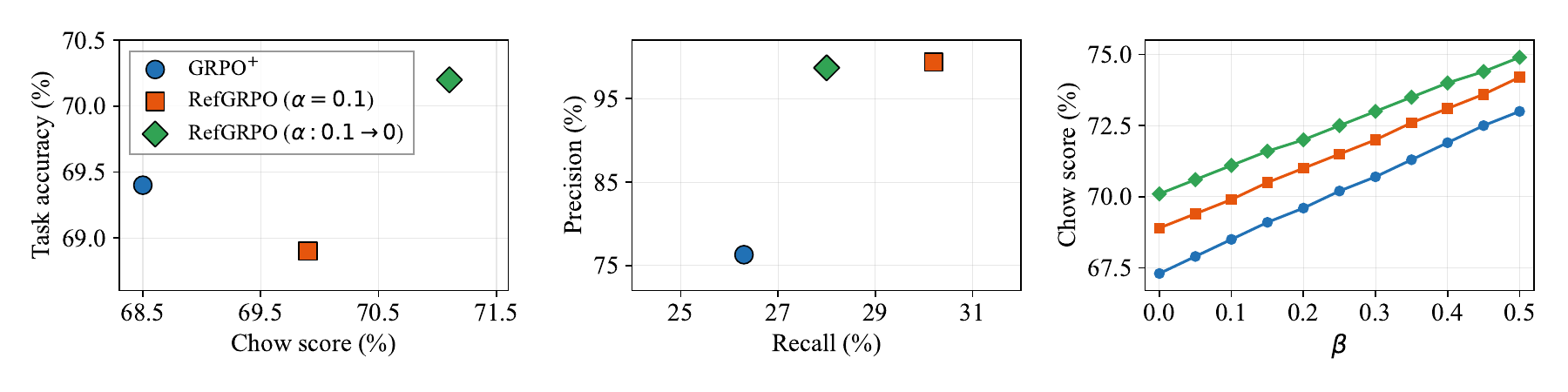}
  \caption{
  Comparison of \grpoplus,
  \myalgo with a fixed schedule $\refcoeff=0.1$, and \myalgo with a dynamic
  schedule $\refcoeff:0.1\to  0$ in the single-turn setting using Qwen2.5-Coder-3B-Instruct as the base model (averaged across 5 benchmarks). \textbf{(a)} Task accuracy
   vs. \chowtext at $\beta= 0.1$. \textbf{(b)} Precision
  vs. recall for error detection. \textbf{(c)} \chowtext across
  error-flag credits $\beta \in [0, 0.5]$.}
  \label{fig:ablation}
\end{figure}

\paragraph{Dynamic vs.\ fixed schedule.}
\cref{fig:ablation}(a) plots \acctext vs. \chowtext for \grpoplus,
\myalgo with a fixed schedule $\refcoeff=0.1$, and \myalgo with a dynamic
schedule $\refcoeff:0.1\to  0$. \myalgo with a fixed schedule improves
\chowtext over \grpoplus but at the cost of a small drop in \acctext; \myalgo with a 
dynamic schedule front-loads calibration and then switches to optimizing task performance, recovering the drop in \acctext and dominating \grpoplus on both metrics.

\paragraph{Error detection analysis.}
\cref{fig:ablation}(b) shows the precision-recall trade-off for error
detection. \grpoplus has moderate precision ($76.3$) and low recall
($26.3$). Both \myalgo variants
reach precision $\geq 98.7$. The variant with a fixed schedule $\refcoeff = 0.1$ achieves the highest
recall ($30.2$) while the variant with a dynamic schedule $\refcoeff: 0.1 \to 0$ accepts a small recall drop in exchange for higher
\acctext (as shown in \cref{fig:ablation}(a)).

\paragraph{\chowtext across error-flag credits.}
\cref{fig:ablation}(c) plots $\chowmath_\beta$ across error-flag credits
$\beta \in [0, 0.5]$. \myalgo dominates \grpoplus across the entire range: 
\myalgo with a 
dynamic schedule $\refcoeff : 0.1 \to 0$ leads by $+2.8$ at $\beta=0$ and $+1.9$ at
$\beta=0.5$. 
This demonstrates that the 
improvements in $\chowmath_\beta$ induced by \myalgo are robust 
to the choice of $\beta$.

\paragraph{Calibration coefficient sweep (multi-turn).}
\Cref{tab:coeff_sweep} sweeps the calibration coefficient with fixed schedules
$\refcoeff \in \{0, 0.1, 0.2\}$ and compares against
the variant with a dynamic schedule ($\refcoeff: 0.1 \to 0$) in the multi-turn setting. Increasing $\refcoeff$
improves \refacctext\ monotonically (from $75.8$ at $\refcoeff = 0$ to  $76.5$ at $\refcoeff = 0.1$, and to $77.4$ at $\refcoeff = 0.2$) but can degrade
\acctext\ (e.g., from $72.5$ at $\refcoeff=0.1$ to  $71.5$ at $\refcoeff=0.2$); the dynamic schedule
$\refcoeff: 0.1\to0$ attains the highest
\acctext\ and \chowtext\ simultaneously.

\begin{table}[t]
\centering
\caption{
  Effect of the calibration coefficient schedule for \myalgo in the multi-turn setting using Qwen2.5-Coder-3B-Instruct as the base model: fixed schedules $\refcoeff \in \{0, 0.1, 0.2\}$ versus a dynamic schedule $\refcoeff: 0.1 \to 0$. Note that $\refcoeff = 0$ recovers \grpoplus.
All values are percentages. Best results are highlighted in \textbf{bold}. Results are averaged across 5 benchmarks.}
\label{tab:coeff_sweep}
\small
\setlength{\tabcolsep}{6pt}
\begin{tabular}{lccc}
\toprule
Schedule & $\accmath$ $\uparrow$ & $\refaccmath$ $\uparrow$
  & $\chowmath_{0.1}$ $\uparrow$ \\
\midrule
$\refcoeff = 0$ (\grpoplus)         & 72.5          & 75.8          & 70.8 \\
$\refcoeff = 0.1$ (fixed)           & 72.5          & 76.5          & 72.6 \\
$\refcoeff = 0.2$ (fixed)           & 71.5          & \textbf{77.4} & 71.9 \\
$\refcoeff : 0.1 \to 0$ (dynamic) & \textbf{72.9} & 76.3 & \textbf{73.1} \\
\bottomrule
\end{tabular}
\end{table}

%% file: sections/related.tex
\section{Related Work}
\label{sec:related}

\paragraph{Reinforcement learning for LLMs.}
RL has become a central post-training paradigm for large language
models, driving substantial gains across diverse
domains \citep{ouyang2022training, bai2022constitutional,
achiam2023gpt, comanici2025gemini}. Progress is especially
pronounced in domains with verifiable rewards such as math and
coding, where the correctness of each action provides a clean
training signal \citep{shao2024deepseekmath, jaech2024openai,
guo2025deepseek, yu2025dapo, liu2025understanding, zheng2025group,
he2025justrl}. 
More recently, attention has
shifted to agentic settings where models interact with environments
over multiple turns and tool calls \citep{jin2025searchr,
wang2025ragen, cao2025skyrl, wei2025reinforcing,
gandhi2026endless, zhang2026the}.
However, the primary objective of these works 
remains task accuracy alone. 
This includes concurrent and independent work \citep{shrivastava2026echo}, which also learns from environment feedback but optimizes a cross-entropy loss on observation tokens rather than training the agent's reflection to be calibrated.
We instead develop an RL algorithm
that, beyond task accuracy, trains the agent to produce calibrated reflection on environment
feedback: with a free calibration bonus and a dynamic schedule on the calibration coefficient,
our method simultaneously improves reflection calibration and task performance relative
to standard outcome-only RL.

\paragraph{Selective prediction and calibration.}
Selective prediction with an abstention option, and the associated
error-reject tradeoff, are classical topics in statistical
learning \citep{chow1957optimum, chow1970optimum}: abstaining on
uncertain inputs yields polynomial sample-complexity speedups for
passive learners \citep{bousquet2021fast} and 
exponential speedups for active learners \citep{zhu2022active, zhu2022efficient}.
A closely related concept, confidence calibration, studies how to
align self-predicted confidence with empirical
correctness \citep{guo2017calibration}.
These
ideas have recently been imported into LLM research, where models are
probed for whether they know what they
know \citep{kadavath2022language, tian2023just, xiong2024can} or 
trained to be better 
calibrated \citep{tao2024trust, leng2025taming, bani-harouni2026rewarding}.
Closest to our setting is
\citet{damani2026beyond}, which uses RL with a 
\emph{fixed-coefficient} Brier bonus  
on the model's confidence in its
self-generated answer \emph{without environment feedback}. 
We differ in two key respects. First, our agent
reflects on \emph{environment feedback}---execution results or 
error messages---in the agentic setting, a strictly richer
signal than self-assessment over a self-generated answer. 
Second, we introduce a \emph{dynamic} schedule on the calibration coefficient,
which is important for simultaneously improving calibration and task accuracy: \cref{fig:ablation}(a) shows that a fixed
coefficient improves calibration at the cost of degrading task accuracy.
\looseness=-1

\paragraph{Self-improvement in LLMs.}
As LLMs are increasingly deployed as autonomous agents, a central
question is how they can improve themselves over time. One line of
work performs iterative verbal self-refinement at inference time,
guided by self-generated critiques (e.g., via prompting) or external
feedback \citep{shinn2023reflexion, madaan2023self, gou2024critic}. 
A complementary line uses gradient-based updates
(SFT or RL) to train models to self-correct or self-improve,
optimizing for a more correct final answer with self-generated or external signals \citep{kumar2025training, ma2025s2r,
zuo2025ttrl, zhu2026testtime}.
Despite some promising
results, \citet{huang2024large} find that LLMs can struggle to
self-correct solely based on self-generated signals.
Our approach differs in \emph{what} it trains. Rather than training the
policy to directly produce more correct answers, we train the agent's
\emph{reflection} to \emph{accurately judge} its own correctness after
observing environment feedback---a calibration objective orthogonal to
task performance. This yields the reliable signal that naive
self-correction lacks: once calibrated, the reflections serve as 
pseudo-rewards for further self-improvement without outcome supervision.
In effect, our method turns the agent into its own verifier grounded in environment feedback, instead of requiring separately trained verifiers \citep{cobbe2021training}.
\looseness=-1

%% file: sections/conclusion.tex
\section{Discussion}
\label{sec:conclusion}

We identified a persistent reflection gap: LLM agents mis-assess their own actions even after observing concrete environment feedback, and outcome-only RL barely fixes it. We offer a simple yet effective fix that augments the outcome reward with a calibration bonus and a dynamic schedule on its coefficient. The calibration bonus is computed for free by contrasting the agent's own reflection with the actual outcome, while the dynamic schedule enables the agent to simultaneously improve reflection calibration (e.g., reduces underconfidence rate $44.4\% \to 7.7\%$) and task accuracy (e.g., $75.1\% \to 76.5\%$), lifting the unified \chowtext metric $73.0\% \to 76.5\%$. 
The resulting calibrated reflection turns the agent into its own verifier grounded in environment feedback, which further enables (i) better self-improvement that uses reflections as pseudo-rewards without outcome supervision, and (ii) more effective test-time selective prediction by committing only to rollouts flagged as correct.
\looseness=-1

\paragraph{Limitations and future work.} Due to compute constraints, the largest model we trained is at the 7B scale; validating \myalgo at larger scales is an important next step. In addition, we focus on a binary reflection score $\refsc \in \{0, 1\}$, and extending it to real-valued confidence scores in $[0,1]$ is an interesting direction.
\looseness=-1

%% file: sections/appendix.tex
\section{Additional Experimental Details and Results}
\label{app:additional_results}

\subsection{Implementation Details}
\label{app:implementation}

\Cref{tab:hyperparams} lists the training hyperparameters shared
between \grpoplus and \myalgo. The maximum response length is
enforced \emph{per turn}, so a $k$-turn rollouts can emit up to
$3000 \times k$ tokens in total. 

\begin{table}[htbp]
\centering
\caption{Training hyperparameters shared by \grpoplus and \myalgo.
}
\label{tab:hyperparams}
\small
\begin{tabular}{ll}
\toprule
Hyperparameter & Value \\
\midrule
Train Batch Size                  & 128 \\
Max Response Length \emph{per turn} & 3000 \\
$(\epslow, \epshigh)$& $(0.2, 0.28)$ \\
Optimizer                         & AdamW \\
Learning Rate                     & 1$\times 10^{-6}$ (constant) \\
Weight Decay                       & 1$\times 10^{-2}$ \\
Temperature                       & 1.0 \\
Rollout $N$                       & 8 \\
\bottomrule
\end{tabular}
\end{table}

For evaluation, we download all base and external models from Hugging Face and
run our own evaluation. The external models (OmniSQL-7B and SQL-R1-7B) are
evaluated only in the single-turn setting, since they are trained in a
single-turn format with long chain-of-thought reasoning.
For reflection accuracy (\refacctext), we report the conditional variant, which
normalizes over the samples in which the model actually produces a valid
reflection.\footnote{The other metrics are unaffected by this normalization:
task accuracy does not depend on reflection; the overconfidence
rate and underconfidence rate are themselves conditional
probabilities; and for the Chow score, a sample with an invalid reflection score
is scored by its task accuracy rather than discarded.} \Cref{tab:valid_reflection}
reports the probability of generating such a valid reflection: it is consistently
lower for the base and external models, whereas our trained models (\grpoplus and
\myalgo) are almost always $100\%$ (and $99.9\%$ in the rare exception). Crucially, although this conditional normalization
\emph{inflates} \refacctext for the base and external models---since samples with
invalid reflections are not counted as errors---\myalgo still attains the best
average \refacctext overall (as shown in \cref{tab:main_single,tab:main_multi}).

\begin{table}[H]
\centering
\caption{Probability of generating a valid reflection (\validtext) for the base
and external models. We omit our trained models
(\grpoplus and \myalgo), whose \validtext is almost always $100\%$ (and $99.9\%$ in the rare exception).
All values are percentages (\emph{higher is better}). Results are averaged across 5 benchmarks.}
\label{tab:valid_reflection}
\small
\setlength{\tabcolsep}{5pt}
\begin{tabular}{lcccccc}
\toprule
 & \multicolumn{2}{c}{Base models: single-turn} & \multicolumn{2}{c}{Base models: multi-turn}
   & \multicolumn{2}{c}{External 7B specialists: single-turn} \\
\cmidrule(lr){2-3} \cmidrule(lr){4-5} \cmidrule(lr){6-7}
 & Llama-3B & Qwen-3B & Qwen-3B & Qwen-7B & OmniSQL-7B & SQL-R1-7B \\
\midrule
$\validmath$ & 89.0 & 98.6 & 87.6 & 85.1 & 95.4 & 97.6 \\
\bottomrule
\end{tabular}
\end{table}

\subsection{Per-Dataset Results for the Main Results}
\label{sec:appendix:per_dataset}

\cref{tab:per_dataset_single,tab:per_dataset_multi} provide per-dataset results for the main results presented in \cref{sec:exp:main}.

\begin{table}[H]
\centering
\caption{Per-dataset single-turn results across five metrics: task accuracy, reflection accuracy, overconfidence rate, underconfidence rate, and Chow score. All values are percentages. Each block reports one metric across the five benchmarks plus the average; arrows in the block headers indicate the desired direction. Best per-dataset results within each scale block are highlighted in \textbf{bold}.
\looseness=-1
}
\label{tab:per_dataset_single}
\small
\setlength{\tabcolsep}{4pt}
\begin{tabular}{llcccccc}
\toprule
Base & Method & Spider-Dev & Spider-DK & Spider-Realistic & Spider-Test & Bird-Dev & Avg. \\
\midrule
\multicolumn{8}{l}{\emph{Task accuracy} $\accmath$ $\uparrow$} \\
\multirow{3}{*}{Llama-3B} & Base       & 63.1 & 53.1 & 57.7 & 60.5 & 26.1 & 52.1 \\
                          & \grpoplus  & 74.6 & 63.4 & \textbf{71.1} & 72.1 & 33.1 & 62.9 \\
                          & \myalgo (Ours) & \textbf{76.4} & \textbf{63.6} & 70.9 & \textbf{74.3} & \textbf{35.0} & \textbf{64.0} \\
\cmidrule(lr){1-8}
\multirow{3}{*}{Qwen-3B}  & Base       & 65.2 & 57.4 & 60.0 & 66.1 & 35.2 & 56.8 \\
                          & \grpoplus  & 79.2 & 66.9 & \textbf{75.0} & 78.7 & \textbf{47.4} & 69.4 \\
                          & \myalgo (Ours) & \textbf{80.6} & \textbf{69.9} & \textbf{75.0} & \textbf{79.6} & 45.9 & \textbf{70.2} \\
\midrule
\multicolumn{8}{l}{\emph{Reflection accuracy} $\refaccmath$ $\uparrow$} \\
\multirow{3}{*}{Llama-3B} & Base       & 68.4 & 58.7 & 68.4 & 69.1 & 60.9 & 65.1 \\
                          & \grpoplus  & 77.2 & 67.5 & 75.8 & 80.5 & 71.3 & 74.5 \\
                          & \myalgo (Ours) & \textbf{79.4} & \textbf{68.6} & \textbf{77.6} & \textbf{81.9} & \textbf{72.7} & \textbf{76.0} \\
\cmidrule(lr){1-8}
\multirow{3}{*}{Qwen-3B}  & Base       & 70.9 & 64.1 & 70.2 & 76.4 & \textbf{75.5} & 71.4 \\
                          & \grpoplus  & 81.0 & 69.3 & 78.7 & 81.8 & 72.3 & 76.6 \\
                          & \myalgo (Ours) & \textbf{85.3} & \textbf{79.6} & \textbf{80.3} & \textbf{82.8} & 70.3 & \textbf{79.7} \\
\midrule
\multicolumn{8}{l}{\emph{Overconfidence rate} $\overmath$ $\downarrow$} \\
\multirow{3}{*}{Llama-3B} & Base       & 30.0 & 39.7 & 32.1 & 30.8 & 55.1 & 37.5 \\
                          & \grpoplus  & 18.7 & \textbf{29.2} & \textbf{19.6} & 19.4 & 45.8 & 26.5 \\
                          & \myalgo (Ours) & \textbf{18.4} & 29.8 & 20.6 & \textbf{18.9} & \textbf{43.8} & \textbf{26.3} \\
\cmidrule(lr){1-8}
\multirow{3}{*}{Qwen-3B}  & Base       & 27.7 & 36.3 & 29.5 & 24.2 & 39.1 & 31.4 \\
                          & \grpoplus  & 17.9 & 29.1 & 20.9 & 18.3 & \textbf{36.7} & 24.6 \\
                          & \myalgo (Ours) & \textbf{15.4} & \textbf{22.6} & \textbf{20.8} & \textbf{17.6} & 39.2 & \textbf{23.1} \\
\midrule
\multicolumn{8}{l}{\emph{Underconfidence rate} $\undermath$ $\downarrow$} \\
\multirow{3}{*}{Llama-3B} & Base       & 41.3 & 47.9 & \textbf{28.9} & 31.5 & 7.3 & 31.4 \\
                          & \grpoplus  & 42.5 & 42.0 & 39.7 & 20.4 & 9.9 & 30.9 \\
                          & \myalgo (Ours) & \textbf{37.2} & \textbf{37.6} & 31.1 & \textbf{11.0} & \textbf{0.5} & \textbf{23.5} \\
\cmidrule(lr){1-8}
\multirow{3}{*}{Qwen-3B}  & Base       & 35.1 & 34.5 & 30.8 & 20.3 & 5.5 & 25.2 \\
                          & \grpoplus  & 34.9 & 41.4 & 25.6 & 15.3 & 1.5 & 23.7 \\
                          & \myalgo (Ours) & \textbf{0.0} & \textbf{0.0} & \textbf{0.0} & \textbf{5.3} & \textbf{1.0} & \textbf{1.3} \\
\midrule
\multicolumn{8}{l}{\emph{Chow score at $\beta = 0.1$} $\chowmath_{0.1}$ $\uparrow$} \\
\multirow{3}{*}{Llama-3B} & Base       & 58.9 & 46.4 & 54.8 & 57.9 & 26.8 & 49.0 \\
                          & \grpoplus  & 68.9 & 55.1 & 64.3 & 70.6 & 33.2 & 58.4 \\
                          & \myalgo (Ours) & \textbf{73.2} & \textbf{57.9} & \textbf{67.1} & \textbf{74.2} & \textbf{38.6} & \textbf{62.2} \\
\cmidrule(lr){1-8}
\multirow{3}{*}{Qwen-3B}  & Base       & 60.6 & 52.3 & 54.7 & 64.4 & 37.1 & 53.8 \\
                          & \grpoplus  & 77.7 & 62.8 & 73.8 & 78.4 & \textbf{49.6} & 68.5 \\
                          & \myalgo (Ours) & \textbf{81.0} & \textbf{70.9} & \textbf{75.5} & \textbf{79.8} & 48.1 & \textbf{71.1} \\
\bottomrule
\end{tabular}
\end{table}

\begin{table}[H]
\centering
\caption{Per-dataset multi-turn results across five metrics: task accuracy, reflection accuracy, overconfidence rate, underconfidence rate, and Chow score. All values are percentages. Each block reports one metric across the five benchmarks plus the average; arrows in the block headers indicate the desired direction. Best per-dataset results within each scale block are highlighted in \textbf{bold}.
\looseness=-1
}
\label{tab:per_dataset_multi}
\small
\setlength{\tabcolsep}{4pt}
\begin{tabular}{llcccccc}
\toprule
Base & Method & Spider-Dev & Spider-DK & Spider-Realistic & Spider-Test & Bird-Dev & Avg. \\
\midrule
\multicolumn{8}{l}{\emph{Task accuracy} $\accmath$ $\uparrow$} \\
\multirow{3}{*}{Qwen-3B} & Base       & 63.4 & 55.7 & 57.7 & 66.1 & 43.0 & 57.2 \\
                    & \grpoplus  & \textbf{81.5} & \textbf{72.0} & 76.0 & 81.0 & 52.2 & 72.5 \\
                    & \myalgo (Ours) & \textbf{81.5} & 71.8 & \textbf{77.0} & \textbf{81.1} & \textbf{52.9} & \textbf{72.9} \\
\cmidrule(lr){1-8}
\multirow{3}{*}{Qwen-7B} & Base       & 75.0 & 63.6 & 74.6 & 76.9 & 53.3 & 68.7 \\
                    & \grpoplus  & 83.3 & 70.8 & 79.3 & 83.3 & 58.7 & 75.1 \\
                    & \myalgo (Ours) & \textbf{83.7} & \textbf{72.7} & \textbf{80.9} & \textbf{83.8} & \textbf{61.3} & \textbf{76.5} \\
\midrule
\multicolumn{8}{l}{\emph{Reflection accuracy} $\refaccmath$ $\uparrow$} \\
\multirow{3}{*}{Qwen-3B} & Base       & 70.9 & 69.2 & 67.3 & 77.2 & \textbf{68.8} & 70.7 \\
                    & \grpoplus  & 81.9 & 72.3 & 75.6 & 82.0 & 67.1 & 75.8 \\
                    & \myalgo (Ours) & \textbf{83.1} & \textbf{73.8} & \textbf{78.7} & \textbf{82.4} & 63.3 & \textbf{76.3} \\
\cmidrule(lr){1-8}
\multirow{3}{*}{Qwen-7B} & Base       & 76.2 & 66.2 & 78.3 & 77.6 & \textbf{67.3} & 73.1 \\
                    & \grpoplus  & 82.3 & 70.3 & 78.5 & 83.7 & 66.3 & 76.2 \\
                    & \myalgo (Ours) & \textbf{84.6} & \textbf{74.2} & \textbf{81.9} & \textbf{84.5} & 64.0 & \textbf{77.8} \\
\midrule
\multicolumn{8}{l}{\emph{Overconfidence} $\overmath$ $\downarrow$} \\
\multirow{3}{*}{Qwen-3B} & Base       & 29.9 & 34.0 & 34.6 & 24.5 & 41.1 & 32.8 \\
                    & \grpoplus  & \textbf{16.6} & \textbf{25.8} & \textbf{21.4} & \textbf{17.6} & \textbf{38.6} & \textbf{24.0} \\
                    & \myalgo (Ours) & 17.1 & 26.7 & 21.6 & 17.8 & 40.9 & 24.8 \\
\cmidrule(lr){1-8}
\multirow{3}{*}{Qwen-7B} & Base       & 21.0 & 30.8 & 18.5 & 20.8 & \textbf{33.7} & 25.0 \\
                    & \grpoplus  & \textbf{15.4} & 27.3 & 19.2 & 15.9 & 36.0 & 22.8 \\
                    & \myalgo (Ours) & 15.5 & \textbf{26.1} & \textbf{18.2} & \textbf{15.6} & 36.8 & \textbf{22.4} \\
\midrule
\multicolumn{8}{l}{\emph{Underconfidence} $\undermath$ $\downarrow$} \\
\multirow{3}{*}{Qwen-3B} & Base       & 24.8 & 21.7 & 25.0 & 13.1 & 8.8 & 18.7 \\
                    & \grpoplus  & 46.2 & 47.8 & 52.0 & 33.9 & 2.5 & 36.5 \\
                    & \myalgo (Ours) & \textbf{5.6} & \textbf{0.0} & \textbf{0.0} & \textbf{3.2} & \textbf{2.4} & \textbf{2.2} \\
\cmidrule(lr){1-8}
\multirow{3}{*}{Qwen-7B} & Base       & 69.2 & 56.9 & 64.5 & 59.8 & 21.3 & 54.3 \\
                    & \grpoplus  & 59.3 & 52.9 & 56.7 & 40.0 & \textbf{13.3} & 44.4 \\
                    & \myalgo (Ours) & \textbf{0.0} & \textbf{10.0} & \textbf{14.3} & \textbf{0.0} & 14.0 & \textbf{7.7} \\
\midrule
\multicolumn{8}{l}{\emph{Chow score at $\beta = 0.1$} $\chowmath_{0.1}$ $\uparrow$} \\
\multirow{3}{*}{Qwen-3B} & Base       & 61.4 & 53.1 & 55.2 & 65.8 & 43.3 & 55.8 \\
                    & \grpoplus  & 79.7 & 68.7 & 71.9 & 80.4 & 53.4 & 70.8 \\
                    & \myalgo (Ours) & \textbf{81.6} & \textbf{72.0} & \textbf{77.1} & \textbf{81.2} & \textbf{53.7} & \textbf{73.1} \\
\cmidrule(lr){1-8}
\multirow{3}{*}{Qwen-7B} & Base       & 72.1 & 59.1 & 71.3 & 75.0 & 52.6 & 66.0 \\
                    & \grpoplus  & 80.7 & 66.7 & 76.6 & 82.7 & 58.4 & 73.0 \\
                    & \myalgo (Ours) & \textbf{83.8} & \textbf{72.7} & \textbf{80.8} & \textbf{83.9} & \textbf{61.2} & \textbf{76.5} \\
\bottomrule
\end{tabular}
\end{table}

\subsection{Per-Dataset External 7B Comparison}
\label{sec:appendix:per_dataset_external}

\Cref{tab:per_dataset_external} reports per-dataset results for the two external 7B SQL specialists discussed in
\cref{fig:external_comparison}.

\begin{table}[h]
\centering
\caption{Per-dataset results for two external 7B SQL specialists (OmniSQL-7B, SQL-R1-7B) across two metrics: task accuracy and reflection accuracy. All values are percentages. Per-dataset results for our \myalgo 7B are reported in \cref{tab:per_dataset_multi}.}
\label{tab:per_dataset_external}
\small
\setlength{\tabcolsep}{4pt}
\begin{tabular}{llcccccc}
\toprule
Model & Metric & Spider-Dev & Spider-DK & Spider-Realistic & Spider-Test & Bird-Dev & Avg. \\
\midrule
\multirow{2}{*}{OmniSQL-7B}
  & $\accmath$               & 82.1 & 72.1 & 77.0 & 82.6 & 63.2 & 75.4 \\
  & $\refaccmath$            & 80.3 & 72.4 & 74.8 & 79.7 & 64.8 & 74.4 \\
\midrule
\multirow{2}{*}{SQL-R1-7B}
  & $\accmath$               & 82.9 & 71.8 & 79.5 & 83.6 & 64.1 & 76.4 \\
  & $\refaccmath$            & 82.4 & 72.2 & 80.2 & 82.9 & 65.4 & 76.6 \\
\bottomrule
\end{tabular}
\end{table}

\subsection{Per-Dataset Self-Improvement Results}
\label{sec:appendix:per_dataset_self}

\Cref{tab:per_dataset_self} reports per-dataset results for the self-improvement results presented in \cref{sec:exp:self}.

\begin{table}[H]
\centering
\caption{Per-dataset task accuracy before ($\accmath_\mathrm{Ckpt}$) and after ($\accmath_\mathrm{Self}$) self-improvement in the single-turn setting using Qwen2.5-Coder-3B-Instruct as the base model. $\Delta = \accmath_\mathrm{Self} - \accmath_\mathrm{Ckpt}$ denotes the change from checkpoint to self-improvement. All values are percentages. Best
per-dataset $\Delta$ is highlighted in \textbf{bold}.}
\label{tab:per_dataset_self}
\small
\setlength{\tabcolsep}{4pt}
\begin{tabular}{lcccccc}
\toprule
 & \multicolumn{3}{c}{\grpoplus} & \multicolumn{3}{c}{\myalgo (Ours)} \\
\cmidrule(lr){2-4} \cmidrule(lr){5-7}
Dataset & $\accmath_\mathrm{Ckpt}$ & $\accmath_\mathrm{Self}$ & $\Delta$ & $\accmath_\mathrm{Ckpt}$ & $\accmath_\mathrm{Self}$ & $\Delta$ \\
\midrule
Spider-Dev   & 78.0 & 77.9 & $-0.1$ & 77.4 & 78.4 & $\mathbf{+1.0}$ \\
Spider-DK    & 66.5 & 70.3 & $\mathbf{+3.8}$ & 65.4 & 69.0 & $+3.6$ \\
Spider-Realistic & 74.4 & 71.5 & $-2.9$ & 72.4 & 74.4 & $\mathbf{+2.0}$ \\
Spider-Test  & 78.5 & 78.7 & $+0.2$ & 78.3 & 79.6 & $\mathbf{+1.3}$ \\
Bird-Dev     & 44.3 & 45.4 & $+1.1$ & 41.9 & 48.2 & $\mathbf{+6.3}$ \\
\midrule
Average      & 68.3 & 68.8 & $+0.5$ & 67.1 & 69.9 & $\mathbf{+2.8}$ \\
\bottomrule
\end{tabular}
\end{table}

\subsection{Per-Dataset Selective Prediction Results}
\label{sec:appendix:per_dataset_select}

\cref{tab:per_dataset_select_all} provides per-dataset results for the selective prediction results presented in \cref{sec:exp:select}.

  \begin{table}[H]
    \centering
    \caption{Per-dataset selective-prediction results: average pass
    rate ($\mathrm{Avg}@8$), accuracy of selective prediction
    ($\selaccmath@8$), and the gain $\Delta = \selaccmath@8 - \mathrm{Avg}@8$
    from selective prediction. 
    Evaluation protocol follows
\cref{sec:exp:select}, and both single-turn and multi-turn evaluations use 
    Qwen2.5-Coder-3B-Instruct as the base model.
    All values are percentages.
    Best per-dataset $\Delta$ between the two methods is highlighted in
    \textbf{bold}.}
    \label{tab:per_dataset_select_all}
    \small
    \setlength{\tabcolsep}{4pt}
    \begin{tabular}{lcccccc}
    \toprule
     & \multicolumn{3}{c}{\grpoplus} & \multicolumn{3}{c}{\myalgo (Ours)} \\
    \cmidrule(lr){2-4} \cmidrule(lr){5-7}
    Dataset & $\mathrm{Avg}@8$ & $\selaccmath@8$ & $\Delta$ & $\mathrm{Avg}@8$ & $\selaccmath@8$ & $\Delta$ \\
    \midrule
    \multicolumn{7}{c}{\textsc{Single-turn}} \\
    \midrule
    Spider-Dev       & 79.3 & 79.1 & $-0.2$          & 80.5 & 81.7 & $\mathbf{+1.2}$ \\
    Spider-DK        & 67.0 & 65.2 & $-1.8$          & 69.9 & 71.2 & $\mathbf{+1.3}$ \\
    Spider-Realistic & 74.3 & 74.7 & $+0.4$          & 75.3 & 76.9 & $\mathbf{+1.6}$ \\
    Spider-Test      & 79.0 & 79.7 & $+0.7$          & 79.8 & 80.9 & $\mathbf{+1.1}$ \\
    Bird-Dev         & 46.6 & 50.2 & $\mathbf{+3.6}$ & 45.9 & 48.6 & $+2.7$ \\
    \midrule
    Average          & 69.2 & 69.8 & $+0.6$          & 70.3 & 71.9 & $\mathbf{+1.6}$ \\
    \midrule
    \midrule
    \multicolumn{7}{c}{\textsc{Multi-turn}} \\
    \midrule
    Spider-Dev       & 80.7 & 80.9 & $+0.2$          & 81.1 & 82.0 & $\mathbf{+0.9}$ \\
    Spider-DK        & 70.8 & 71.7 & $+0.9$          & 71.1 & 72.3 & $\mathbf{+1.2}$ \\
    Spider-Realistic & 76.6 & 76.7 & $+0.1$          & 76.5 & 77.4 & $\mathbf{+0.9}$ \\
    Spider-Test      & 80.8 & 81.3 & $+0.5$          & 81.1 & 81.7 & $\mathbf{+0.6}$ \\
    Bird-Dev         & 52.7 & 55.5 & $\mathbf{+2.8}$ & 52.7 & 54.8 & $+2.1$ \\
    \midrule
    Average          & 72.3 & 73.2 & $+0.9$          & 72.5 & 73.6 & $\mathbf{+1.1}$ \\
    \bottomrule
    \end{tabular}
    \end{table}